%% file: paper.tex
\documentclass[sigconf]{article}
\usepackage{microtype}
\usepackage{graphicx}
\usepackage{enumitem}
\usepackage{subfigure}
\usepackage{multirow}
\usepackage{float}
\usepackage{enumerate}   
\usepackage{hyperref}
\usepackage{balance}
\usepackage{flushend}
\usepackage{adjustbox}
\usepackage{colortbl}
\usepackage{algorithm} 
\usepackage{algorithmic}  
\usepackage[algo2e]{algorithm2e} 
\usepackage[utf8]{inputenc}
\usepackage[T1]{fontenc}    
\usepackage{multirow}
\usepackage{hyperref}       
\usepackage{url}            
\usepackage{booktabs}       
\usepackage{nicefrac}       
\usepackage{microtype}      
\usepackage{amsthm}
\usepackage{amsmath}
\usepackage{amssymb}
\usepackage{amsfonts}       
\usepackage{mathtools}
\usepackage{mathrsfs}
\usepackage{xcolor}
\usepackage{relsize}    
\usepackage{graphicx}
\usepackage{enumerate}
\usepackage{pgfplots}
\usepackage{filecontents}
\usepackage{multicol}
\usepackage{tikz}
\usepackage{environ}
\usepackage[symbol]{footmisc}
\usepackage{pgfplots}
\usetikzlibrary{3d}
\usepackage{wrapfig}
\usepackage{tikz-3dplot}
\usepgfplotslibrary{patchplots}
\usetikzlibrary{arrows}
\usepgfplotslibrary{fillbetween}
\usepackage{blindtext}
\usepackage{makecell}
\usepackage[font=scriptsize]{caption}

\usepackage{sidecap}

\SetCommentSty{mycommfont}
\SetKwComment{Comment}{$\blacktriangleleft$\ }{$\blacksquare$}
\SetKwComment{tcc}{$\blacktriangledown$\ }{$\blacktriangledown$}
\SetKwInput{KwInput}{Input}                
\SetKwInput{KwOutput}{Output}              
\SetKwInput{KwParameter}{Parameter}              

\pgfplotsset{compat=1.16} 
\newcommand{\executeiffilenewer}[3]{%
\ifnum\pdfstrcmp{\pdffilemoddate{#1}}%
{\pdffilemoddate{#2}}>0%
{\immediate\write18{#3}}\fi%
}

\usepackage[accepted]{icml2023} 
\icmltitlerunning{Enhancing User Experience in On-Device Machine Learning with Gated Compression Layers}

\begin{document}
\twocolumn[
\icmltitle{Enhancing User Experience in On-Device Machine Learning with Gated Compression Layers}
\icmlsetsymbol{equal}{*}
\begin{icmlauthorlist}
\icmlauthor{Haiguang Li}{google}
\icmlauthor{Usama Pervaiz}{google}
\icmlauthor{Joseph Antognini}{google}
\icmlauthor{Michał Matuszak}{google}
\icmlauthor{Lawrence Au}{google}
\icmlauthor{Gilles Roux}{google}
\icmlauthor{Trausti Thormundsson}{google}
\end{icmlauthorlist}
\icmlaffiliation{google}{Google LLC, Mountain View, CA 94043, USA}
\icmlcorrespondingauthor{Haiguang Li}{haiguang@google.com}
\icmlkeywords{Machine Learning, ICML}
\vskip 0.3in
]

\printAffiliationsAndNotice{}
\begin{abstract}
On-device machine learning (ODML) enables powerful edge applications, but power consumption remains a key challenge for resource-constrained devices. To address this, developers often face a trade-off between model accuracy and power consumption, employing either computationally intensive models on high-power cores or pared-down models on low-power cores. Both approaches typically lead to a compromise in user experience (UX). This work focuses on the use of Gated Compression (GC) layer to enhance ODML model performance while conserving power and maximizing cost-efficiency, especially for always-on use cases. GC layers dynamically regulate data flow by selectively gating activations of neurons within the neural network and effectively filtering out non-essential inputs, which reduces power needs without compromising accuracy, and enables more efficient execution on heterogeneous compute cores. These improvements enhance UX through prolonged battery life, improved device responsiveness, and greater user comfort.
In this work, we have integrated GC layers into vision and speech domain models including the transformer-based ViT model.
Our experiments demonstrate theoretical power efficiency gains ranging from 158x to 30,000x for always-on scenarios. This substantial improvement empowers ODML applications with enhanced UX benefits.
\end{abstract}

\section{Introduction}

On-device machine learning (ODML), the practice of running machine learning (ML) algorithms directly on a user's device, has emerged as a promising approach to provide more interactive and responsive user experiences. Without the need for data to travel to cloud-based servers, ODML ensures that sensitive information is processed securely on the device, enhancing user privacy \cite{chen2014diannao}. This local processing capability not only allows devices to operate intelligently while offline but also avoids the costs associated with cloud computing, making ODML an economical choice for users and developers alike. ODML facilitates low-latency inference as data can be processed locally by leveraging hardware acceleration on device (i.e., avoiding round trip to the server) resulting in swift and smooth the user experience \cite{murshed2021machine}. Therefore, ODML application in various devices, such as smartphones, wearables, and IoT devices, can allow for real-time, context-aware, and personalized user interactions. However, deploying ML models on resource-constrained devices presents unique power consumption challenges that directly impact the user experience (UX) \cite{han2015deep}.

The transition to ODML brings forth its own set of challenges, foremost among them is the constraint on computational resources inherent to edge devices. Power consumption in ML models ondevice is influenced by factors such as model size and complexity, computational load, and memory access frequency. These factors lead to UX challenges including reduced battery life, inadequate storage and memory, performance trade-offs, decreased responsiveness, and limited functionality \cite{chen2016eyeriss, gupta2015deep}. Furthermore, the thermal effects of continuous computing can lead to device throttling, thereby diminishing UX \cite{zhou2022play}. To address these challenges and deliver a superior UX, it is essential to develop power-efficient techniques and strategies for ODML.

To understand and solve the power consumption challenges in ODML and their impact on UX, researchers have proposed various techniques to address these challenges, including model compression, hardware-aware optimization, and neural architecture search (NAS) \cite{han2015deep, hinton2015distilling, zhou2017incremental, chen2016eyeriss, tan2019mnasnet, li2024dynamic}. Normally, modern devices utilize dedicated heterogeneous hardware, supporting low-power, lightweight ML models. As user expectations grow, it is vital to provide better UX for context-aware models that activate selectively, ensuring that devices run only when needed. This not only provides a better UX by preserving battery life but also aligns with environmentally sustainable computing practices.

Always-on models are a category of ODML models that run continually, searching for potential signals of interest (referred to as positive samples) amidst a steady flow of mostly irrelevant information (referred to as negative samples). Positive samples are less frequent, for example, a keyword in a conversation. Because these models are continually invoked, they can consume a significant amount of power on an edge device. The efficient approach for deploying always-on models would be to consume minimal power resources on negative samples while on a look-out for positive events though those events are sparse.

As ODML models grow in complexity, efficiently using a device's heterogeneous computing cores (e.g., always-on accelerators, DSPs and neural processing units) becomes crucial for energy-efficient execution. In this paper, we deploy larger models with integrated Gated Compression (GC) layer \cite{li2023gcl} that allows for the nuanced use of these heterogeneous cores resulting in better UX. The integration of the GC layer can allow for the early layers of the neural network to operate on ultra-low-power accelerators, focusing on the detection of signal of interest (i.e., positive samples). Subsequently, the later stages, responsible for more complex analyses, are activated only when necessary, and can operate on compute-intensive processors as needed. This tiered computational strategy improves feature detection and system responsiveness without compromising battery life, ultimately enhancing the user experience.

In this study, we aim to improve user experience in ODML by incorporating the Gated Compression Layer \cite{li2023gcl} into pre-existing baseline models, such as ResNet-152 \cite{he2016deep-resnet-152} and TC-ResNet \cite{choi2019temporal-tc-resnet}, which cater to vision and speech domains. Our comprehensive experimental findings indicate that a significant reduction in power consumption can be achieved by efficiently stopping negative samples early and promoting activation sparsity (i.e., compressing intermediate feature data to minimize the data that is propagated through
an active network) for positive samples, leading to an enhanced user experience. 
 
\section{GC Layers for Always-On Models}

The GC layer, as described in \citet{li2023gcl}, is a gating mechanism specifically designed to filter out irrelevant or redundant information within input data. This is achieved by applying a trainable gate function to each neuron's output. The GC layer can be incorporated into existing neural network architectures, with its position optimized to strike a balance between performance and power efficiency (see Figure \ref{fig:gc_4_ux}).

\input{figs/gc4ux.tikz}

In always-on use cases that involve ML models on low-power computing cores, the GC layer contributes to enhanced power efficiency, extended battery life, and better resource utilization, all while maintaining or even improving model accuracy. This is accomplished by reducing data transmission and computation requirements as explained in the following sections.

\subsection{Distributed Model with GC Layers}
Always-on use cases commonly feature multiple heterogeneous compute islands, including sensors, micro controllers, sensor hubs, mobile devices, and even the cloud. GC layers partition existing networks into smaller sub-networks, which can be executed on separate compute islands. GC layers can enhance distributed models by enabling selective activation of network components, ensuring that only the necessary parts of the model are running at any given time. This approach allows for the complete utilization of all available resources, resulting in larger, more powerful networks with enhanced performance.

\subsection{Early Stopping for Negative Samples}
Early-stopping plays a vital role in reducing power consumption and computational resources, while maintaining or even enhancing the model's performance.
Early stopping is determined by monitoring the confidence scores or activation values of the gates inside the GC layers in the neural network during the inference process. If the confidence score for a particular sample surpasses a predefined threshold, the processing is stopped early without invoking the remaining sub networks. Essentially, the model is confident enough in its prediction at this point and does not need to perform further computations through the remainder of the
 network.
\input{figs/dataset_arch_table.tikz}

This early stopping mechanism in the GC layer allows for efficient filtering of negative samples which can be discarded earlier in the processing pipeline. Doing so, it saves power and computation resources as the network does not need to process these samples through all the layers.

\subsection{Activation Sparsity for Positive Samples}
In distributed networks, the GC layer promotes activation sparsity for positive samples; it means only the most crucial connections within the network are activated, corresponding to the most informative features of the input data. In other words, the GC layer induces sparsity within intermediate feature maps of the neural network. This means that during the forward pass, when the network is making a prediction, any computations involving those zeroed feature maps are skipped. This selective activation (akin to feature selection) of feature maps conserves power by reducing the number of computations across the network as we are effectively operating on a reduced subset of feature maps.

In a distributed environment, two adjacent smaller networks operate on distinct compute islands. The output of one sub-network serves as the input to the subsequent sub-network. Data transmission across physical boundaries (e.g., device-to-device communication via Bluetooth or WiFi) consumes significant power. GC layers purposefully establish network bottlenecks to decrease data transmission across boundaries. By limiting the amount of data transferred to the following layers, these bottlenecks lead to reduced power consumption.

\subsection{UX Benefits of GC Layers}

Incorporating GC layers into ODML architectures refines UX by offering smart computation. The early-exit mechanism (i.e., preemptively halting data processing when no relevant signal is present) minimizes unnecessary power usage and computational latency. This can enhance device responsiveness to the user interactions particularly in distributed network environments where computational tasks are split across multiple nodes.

GC layers also enable devices to stay operational for longer periods, conserving energy by avoiding the full activation of feature maps in neural networks (i.e., activation sparsity). This translates to sustained user engagement without frequent interruptions for charging, a critical factor for UX in mobile and wearable technologies.  Additionally, improved power efficiency helps devices operate at cooler temperatures, increasing user comfort during use. Moreover, better power efficiency contributes to enhanced performance, as devices can run more swiftly and seamlessly.

Overall, superior power efficiency contributes to a better UX, as users can enjoy devices that last longer while avoiding needless power drain and thermal buildup. This results in devices that are more convenient, comfortable, and pleasurable to use.

\section{Experiments}
\input{figs/overall_table.tikz}
\input{figs/overall_plot.tikz}

We apply GC layers to always-on scenarios across both the vision and speech domains. In this section, we first describe the datasets, evaluation protocols, and implementation details used to train and test each model, then discuss results. Following this, we explore the outcomes and analyze the different components of the GC layer.

\subsection{Datasets, Architectures, and Implementation Details}

We conduct experiments using ImageNet 2012 \citep{ILSVRC15}, the most common and robust public image dataset, and Speech Command \citep{speechcommandsv2}, a popular audio dataset. For the ImageNet dataset, we tested GC layer performance on two different tasks: person detection and dog detection. For the person detection task, we map all classes without people involved to a generic background class. Similarly, for the dog detection task, we map all classes without dogs to a generic background class while only dog classes are considered as positive classes. This was done to reflect an always-on use case, where data distribution is typically dominated by negative/background classes. The Speech Command dataset already contains a background class with a 1:4 ratio of positive to negative samples, and no additional class-label remapping is needed. It is important to note that real-world tasks often have significant imbalances weighted towards negative samples, highlighting the need for techniques that support early stopping in always-on models. In Table \ref{tab:dataset_and_arch}, we provided detailed information on our datasets and architectures.

We demonstrate that GC layers can be applied to common model architectures by using ResNet-152 \citep{he2016deep-resnet-152} for ImageNet 2012 and TC-ResNet \citep{choi2019temporal-tc-resnet} for Speech Command.

We compare architectures that are expanded with GC layers to baseline architectures that do not include gating and activation sparsity. For each dataset, the GC architecture is identical to the baseline architecture with the exception of an additional GC layer placed at various depths of each network (as shown in Figure \ref{fig:sub_model_size}).

We evaluate the precision, recall, early-stopping, and activation sparsity performance of architectures expanded with GC layers. Early stopping is defined as the percentage of negative test examples that are successfully gated by the model without having to propagate to the final classification layer.

We implemented all methods using TensorFlow 2.x \citep{tensorflow2015-whitepaper}, and used the Adam optimizer \citep{adom2015} with either Cosine Decay or Piecewise Constant Decay learning rate scheduler (as shown in Figure \ref{fig:Learning_Rate}) for model training. All experiments are repeated 10 times with the mean and variance results reported.

\subsection{Model Performance: Precision and Recall}

Table \ref{tab:overall_performance} Compares the baseline models and GC models across different real-world use-cases in the vision and speech domains. Note that GC models consistently achieve better model performance with higher precision and recall compared to the baseline models. 

Figure \ref{fig:imagenet_2012_overall} shows that GCL@3 --- a GC layer placed at 6\% depth of the baseline network --- achieves the highest precision (99.5320\%) and recall (95.7075\%) for ImageNet person detection task.  By comparison, the baseline model without a GC layer achieves a precision of 99.3512\% and a recall of 95.2378\% lower than that of any model with a GC layer.

For the ImageNet dog detection task (Figure \ref{fig:imagenet_2012_overall}), inserting a GC layer also yielded substantial improvements in detection capabilities. For example, GCL@3 --- a GC layer placed at 6\% depth of the baseline network --- achieves the highest precision (97.0122\%) and recall (97.1914\%) for the dog detection task. By contrast, the baseline model achieves a precision of 96.6936\% and recall of 96.27\%, again lower than that of any model with a GC layer.

Similarly, in Figure \ref{fig:speech_command_overall}, GCL@2 --- a GC layer placed at 20\% depth of the baseline network --- yields the highest precision for both the 35 keyword classes detection (96.8182\%) and the 10 keyword classes detection (97.6821\%) on the Speech Command dataset. Additionally, GCL@3 --- a GC layer placed at 30\% depth of the baseline network --- also achieves the highest recall (96.7892\% and 97.8194\%) for 35 keyword and 10 keyword classes detection tasks respectively.

In both the vision and speech domains, the GC models consistently outperform the baseline models in terms of precision and recall. The specific GC layers that achieved the highest precision and recall for different use-cases are highlighted in Figures \ref{fig:imagenet_2012_overall} and \ref{fig:speech_command_overall}. These findings suggest that early layers of neural network with the insertion of GC layer can effectively filter relevant features, enhancing the network's focus and efficiency. Such results underscore the potential of GC layers to refine ODML, bolstering not only computational efficiency but also the accuracy of the models.

\subsection{Gating Performance: Early Stopping}

\input{figs/roc_plots.tikz}

\input{figs/power_costs.tikz}

In this section, we discuss the comparative effectiveness of gating performance as applied to different datasets/models. We also analyze how the GC layer placement impacts it's ability to accurately distinguish negative samples (i.e., background class) from the positive samples (e.g., dog or face detection). For the gating experiments described in this section, we have set a low false negative rate (i.e., incorrect gating rate), ensuring only 1\% of relevant samples (i.e., positive samples) were incorrectly gated. This strict threshold allows us to observe the effectiveness of early stopping by the GC layer, with the correct gating rate illustrating the percentage of irrelevant data (i.e., negative samples) that was successfully halted.

For the ImageNet tasks (person and dog detection), as shown in Figure \ref{fig:imagenet_2012_overall}, the dog detector model outperforms the person detector model in terms of gating performance, achieving 42.3\%, 73.1\%, and 94.9\% early stopping at 2\%, 4\%, and 6\% network depth, respectively. On the other hand, the person detector model shows lower performance, reaching 38.1\%, 45.5\%, and 52.3\% stopping performance at 2\%, 4\%, and 6\% network depth, respectively. This discrepancy in performance can be attributed to the limited number of positive classes (4) for the person detection task in the dataset, in contrast to, 130 positive classes for dog detection task in ImageNet dataset as presented in Table \ref{tab:dataset_and_arch}. In other words, 99.6\% of data is comprised of negative samples for the person detection task in comparison to 86\% of negative samples in dog detection task (as shown in Table \ref{tab:dataset_and_arch}). In the testing set, each class contains merely around 50 samples, leading to a total of merely 200 positive samples for assessment. In order to maintain a low false negative rate ($\leq1\%$), the model can make no more than two mistakes when identifying positive samples. Additionally, the person detector model exhibits higher variance, as each false negative has a more significant impact on the false negative rate. 

For the keyword detection task on Speech Command dataset, as shown in Figure \ref{fig:speech_command_overall}, placing the GC layer between 10\% and 30\% depth of the baseline network consistently leads to 100\% early stopping performance for all GC models. Furthermore, a more detailed receiver operating characteristic plot for each 10 runs in Figure \ref{fig:roc} indicates that the GC layer can accurately identify all background samples without mistakenly predicting any positive samples as negative.

Another observation evident from these results (Figure \ref{fig:imagenet_2012_overall} and Figure \ref{fig:speech_command_overall}) is that the performance of the GC layer significantly improves as it is placed deeper in the network. For example, the performance of early stopping for dog detection task improved from 42.3\% to 94.9\% at 2\% versus 6\% network depth (also demonstrated in Figure \ref{fig:roc} dog detector model results). This improved performance can be attributed to the fact that at greater depths, the GC layer benefits from network's prior computations, which already have filtered much of the noise and less relevant information. Therefore, GC layer can apply a more informed gating strategy, and better identify the signals of interest (i.e., positive samples) with greater precision. 

Early stopping greatly reduces the computational burden by not processing the entire cascade of network layers when it is unnecessary. In doing so, early stopping has a significant impact on improving model performance, accuracy, and efficiency in both vision and speech domain related tasks.

\subsection{Compression Performance: Activation Sparsity}
As illustrated in Figure \ref{fig:imagenet_2012_overall}, the GC models consistently achieve activation sparsity of greater than 98\% and 92\% for the person detection and dog detection tasks, respectively. Likewise, in Figure \ref{fig:speech_command_overall}, the GC models attain activation sparsity ranging from 75\% to 90\% for the 35 keyword classes detection task and from 85\% to 95\% for the 10 keyword classes detection task.

The results presented in Figure \ref{fig:imagenet_2012_overall} and Figure \ref{fig:speech_command_overall} shows that as GC layer is moved to deeper positions, activation sparsity in most cases decreases. This is because, as the GC layer is placed in shallower positions, it can compress more dimensions due to the larger internal feature map size.

Overall, these results demonstrate that the GC layer has consistently demonstrated the ability to achieve high activation sparsity in both the vision and speech domain related tasks.

\subsection{Gating Performance Versus GC Layer's Depth}
When a GC layer is added to an existing network, it creates two sub-networks, with the size of the first sub-model increasing as the position of the GC layer deepens (as illustrated in Figure \ref{fig:sub_model_size}). This larger sub-model allows for better gating performance (as shown in Figure \ref{fig:imagenet_2012_overall}).

Although there is no universally optimal position for placing the GC layer, a deeper position allows for a larger sub-network to be fine-tuned, resulting in better gating performance. However, the optimal position depends on the specific use case and resource constraints. A general approach is to identify the deepest position within resource limits and then evaluate the benefits of moving towards a shallower position.

In terms of the UX benefits, the initial network can reside on a low-power core (i.e., \textit{always-on} core), continuously processing data with minimal energy consumption. When this initial network encounters a positive sample, it triggers the activation of the more powerful core (i.e., \textit{on-demand}). This limits the operations of high-energy cores to only essential tasks. This strategic activation of resources ensures that the device remains responsive and ready for important tasks without unnecessary power drain. 

\subsection{Power Saving: Theoretical Analysis}
In addition to benefits in model performance and size, GC models offer power savings. However, power cost in real-world scenarios is influenced by several variables. In this work, the power cost schema can be simplified into three components: 1) the power cost of running inference on the first sub-network, 2) the power cost of propagating data from the first sub-network to the second sub-network, and 3) the power cost of running inference on the remaining sub-network. Assuming that the power cost of running inference is proportional to the depth of the model with a constant of $a$, and the power cost of propagating data across sub-networks is proportional to the amount of data propagated with a constant of $b$.

For an existing model $\mathcal{M}$, since it requires the computation of the entire network and the propagation of the entire data amount, therefore, the expected power cost of running one inference through can be computed as: 
\begin{equation}
\mathbb{E}_{\text{power}}(\mathcal{M}) = a + b.
\end{equation}
By assigning distinct values to $a$ and $b$, the power expenses of various system configurations can be characterized. A higher value of $a$ implies that executing model inferences requires more power (i.e., computation expensive), whereas a higher value of $b$ implies that transmitting data requires more power (i.e., IO expensive). 

After adding a GC layer, the new GC model $\mathcal{M}'$ is split into two disjoint sub models. Since the GC layer is a simple binary classifier head and a sparse binary mask, its size and computation is small and can be neglected. Then, the expected power cost of running inference on one example can be computed as: 
\begin{equation}\label{eq:power_cost}
\begin{matrix*}[l]
\mathbb{E}&\hspace{-10pt}_{\text{power}}(\mathcal{M}';\rho,\mu,\nu,\gamma) =\\ 
 &\hspace{-8pt}\mu a +  [\rho+(1-\rho)(1-\gamma)][(1-\mu)a+(1-\nu) b],
\end{matrix*}
\end{equation}

\input{figs/vit_table.tikz}
where $\rho$ denotes the probability of selecting a positive sample from the data distribution, $\mu$ represents the ratio of the depth-first sub-network to the entire network, $\nu$ represents the activation sparsity rate, and $\gamma$ denotes the gating rate for negative samples without affecting positive samples.

Since $a+b=1$, we can express $b$ as $b=1-a$. Therefore, given the values of $\rho$, $\mu$, $\nu$, and $\gamma$, the power cost in Equation \ref{eq:power_cost} can be considered as a function of $a$. The derivative of the expected power cost of $\mathcal{M}'$ with respect to $a$ or $b$ can be computed as follows:
\begin{equation}
    \left\{\begin{matrix*}[l]
\frac{\partial
\mathbb{E}_{\text{power}}(\mathcal{M}';\rho,\mu,\nu,\gamma)}{\partial a} &\hspace{-8pt}=\mu +[\rho+(1-\rho)(1-\gamma)](1-\mu)\\&\hspace{-8pt}>0\\
\frac{\partial
\mathbb{E}_{\text{power}}(\mathcal{M}';\rho,\mu,\nu,\gamma)}{\partial b} &\hspace{-8pt}= [\rho+(1-\rho)(1-\gamma)](1-\nu)\\ &\hspace{-8pt}>0
\end{matrix*}\right..
\end{equation}

This indicates that the power cost increases with an increase in the value of $a$ (or $b$). More specifically, for a given use case with fixed $\rho$ and an optimized GC model with fixed $\mu$, $\nu$, and $\gamma$, it is recommended to choose a system configuration with lower computational expense (i.e., lower $a$ value) and transmission expense (i.e., lower $b$ value). Moreover, using the given parameters ($\rho, \mu, \nu$, and $\gamma$), the dividend ratio can be computed to determine the system configuration that offers the best trade-off between computational and IO expense. 

The value of $\rho$ is determined by the data distribution of the use case, which is beyond our control. On the other hand, $\nu$ and $\gamma$ depend on the placement of the GC layer (upper bounded by the placement location), which is controlled by $\mu$. Hence, the crucial factor in adding a GC layer for power efficiency is to select an appropriate placement location.

Figure \ref{fig:power_cost} shows that GC models achieve substantially lower power costs as compared to the original baselines (which require 100\% power cost) across all system configurations, including computation expensive, computation/IO balanced, and IO expensive. Specifically, GC models consume power costs ranging from 0.003\% to 0.63\% of the baseline models, leading to a reduction in power costs by a factor of 158 to 30,000 in comparison to the baseline models.

In summary, results presented in Figure \ref{fig:power_cost} highlights that introduction of GC layer to ODML can introduce a paradigm shift in power consumption patterns, effectively minimizing operational costs without compromising model effectiveness.

\section{Extending GC Layers to Transformers}
While \citet{li2023gcl} and our previous work in Section 3 explored the impact of GC layers on CNN-based models (see models in Table \ref{tab:dataset_and_arch}), we now extend the application of GC layers to the powerful Transformer architecture \cite{vaswani2017attention}. Given their success in NLP \cite{wolf2020transformers} and computer vision \cite{khan2022transformers}, this integration offers exciting opportunities.

Unlike traditional CNNs, Vision Transformers (ViTs) \cite{dosovitskiy2021an} leverage the attention mechanism from NLP transformers to process image data. This novel approach enables them to capture long-range dependencies across an entire image in a way that CNNs cannot due to their local receptive fields. We integrated the GC layer into the ViT-L/16 model.

\input{figs/vit_imagenet_results}

\subsection{Overall Performance}

Table \ref{tab:overall_vit_performance} and Figure \ref{fig:vit-alpha-performance-pdf} summarize the comprehensive performance analysis. Our findings demonstrate consistent improvements in precision and recall compared to the baseline model without GC layers. For example, we increased the precision from the baseline value (i.e., without the GC layer) of 89.448\% to 90.996\% by placing the GC layer at 5\% depth of the baseline network. Notably, we also achieved a remarkable activation sparsity between 89\% to 91\% and early stopping rates between 44\% and 86\%, all while maintaining strong model performance (Figure \ref{fig:vit-alpha-performance-pdf}).

This is particularly significant for ViTs as they are known for their high computational cost, especially with increasing input size and model depth \citep[as cited in][]{mauricio2023comparing}. We believe that the introduction of the GC layer (i.e., activation sparsity and early stopping techniques) has the potential to significantly improve the efficiency of these models without compromising their effectiveness. 

\subsection{Understanding the GC Layer’s Placement}

We investigated how the placement of the GC layer within the network architecture affects its ability to achieve early stopping efficiently. We conducted experiments varying the GC layer's depth within the network at 5\%, 10\%, 25\%, and 50\%. The results, presented in Figure \ref{fig:vit-alpha-performance-pdf}, reveal a clear relationship between GC layer placement and early stopping performance.

As expected, we observed a positive correlation between the base network's size and the achievable early stopping rate (similar to the results of ResNet-152 in the vision domain, Table \ref{tab:overall_performance}). For example, in some cases (Figure \ref{fig:vit-alpha-performance-pdf}, refer to gating performance graph), the gating performance increased from 56\% to 85\% as the GC layer's depth within the network is changed from 5\% to 50\%. This is because placing the GC layer too early in the network limits the processing of features, hindering its ability to make accurate predictions for early stopping. In other words, strategic positioning of the GC layer in a way to allow for  adequate feature computation by the base network is essential for the GC layer to perform accurate early stopping.

\subsection{Understanding the Early Stopping Weight, $\alpha$}
The early stopping weight, $\alpha$, within the GC layer plays a critical role in determining the effectiveness of early stopping. Early stopping offers significant benefits in terms of computational and power efficiency. We conducted a series of experiments to analyze the impact of $\alpha$ on early stopping performance while maintaining the overall model performance.

Figure \ref{fig:vit-alpha-performance-pdf} summarizes the key findings:
(1) Preserved Model Performance: The values of $\alpha$ did not negatively impact model performance metrics such as recall and precision.
(2) Enhanced Early Stopping: A clear positive correlation was observed between $\alpha$ and the achieved early stopping rate. Higher $\alpha$ values resulted in more frequent early stopping.
(3) Modest Impact on Activation Sparsity: As anticipated, there was a slight decrease in activation sparsity with increasing $\alpha$. This is because the activation sparsity weight ($\beta$) is inversely related to $\alpha$, defined as $\beta=1-\alpha$.

These findings indicate that we can effectively use the early stopping weight ($\alpha$) to improve model efficiency through increased early stopping without compromising overall accuracy.

\subsection{Understanding the Incorrect Gating Rate}
High precision and recall performance depends on properly training the GC layer to strike a balance between:
(1) Correct Gating Rate: This metric signifies the GC layer's effectiveness in accurately identifying and stopping irrelevant data (i.e., negative samples).
(2) Incorrect Gating Rate: This metric represents the possibility of the GC layer mistakenly stopping important samples (i.e., positive samples).

In this work, we prioritized minimizing incorrect gating by setting a maximum threshold of 1\% (0.01). However, different use cases may have a different set of trade-offs.

As illustrated in Figure \ref{fig:vit-alpha-performance}, increasing the acceptable incorrect gating rate (while keeping $\alpha$ fixed) presents a trade-off:
(1) Improved Correct Gating: We observe a significant rise in the correct gating rate. For instance, the correct gating rate increased from 0.456 to 0.758 as shown in Figure \ref{fig:vit-alpha-performance} by setting $\alpha$ to 0.1.
(2) Marginal Rise in Incorrect Gating: There is a small upturn in the incorrect gating rate (e.g., from 0.01 to 0.05) suggesting a slight increase in the chance of the GC layer incorrectly interrupting positive samples.

This trade-off empowers customization of the incorrect gating rate based on the specific priorities:
(1) Prioritize High Correct Gating: If maximizing efficiency through accurate stopping is important, tolerating a marginally increased incorrect gating rate may be viable.
(2) Minimize Incorrect Gating: If avoiding any erroneous early exiting of positive samples is critical, we recommend choosing a lower incorrect gating rate.

Understanding this balance allows one to adjust the GC layer for ideal performance, aligning with the specific needs of proposed application.

\section{Conclusion}

This work demonstrates how Gated Compression (GC) layers significantly enhance the power efficiency of on-device machine learning (ODML) models, particularly for always-on use cases. Our approach overcomes the limitations of traditional power-saving methods, enabling power conservation without sacrificing model accuracy. GC layers achieve this through a combination of selective filtering of negative samples and promoting activation sparsity for positive samples within intermediate feature maps of neural networks, minimizing unnecessary computations and data transmission.

The integration of the GC layer into ODML offers substantial user experience benefits, such as by extending battery life, longer uninterrupted usage sessions can be enabled for users. Devices become more responsive with reduced processing latency, enhancing practical utility. Lower power consumption also results in devices operating at cooler temperatures, which adds to user comfort. Additionally, our proposed approach aligns with eco-friendly computing practices.

Our experiments across vision and speech tasks validate the benefits of integrating GC layers. This includes their successful first-time use within the transformer-based ViT model, showcasing their adaptability. In always-on scenarios, we achieved remarkable theoretical power efficiency gains between 158x and 30,000x --- crucially, without sacrificing accuracy. This paves the way for deploying more powerful ODML applications, offering richer features while extending device battery life for a superior user experience.

This research presents exciting directions for future work. Investigating hardware co-design, with hardware-aware optimization of GC layers specifically for accelerators, could maximize power efficiency. Additionally, exploring domain-specific adaptations of GC layers across diverse fields like healthcare and robotics holds significant potential. Ultimately, large-scale real-world testing will illuminate the broader UX impact of GC layers and their role in advancing context-aware computing.

\balance
\bibliography{paper}
\bibliographystyle{icml2023}
\end{document}

%% file: figs/gc4ux.tikz
\begin{figure}[t!]
    \centering
    \includegraphics[width=0.4975\textwidth]{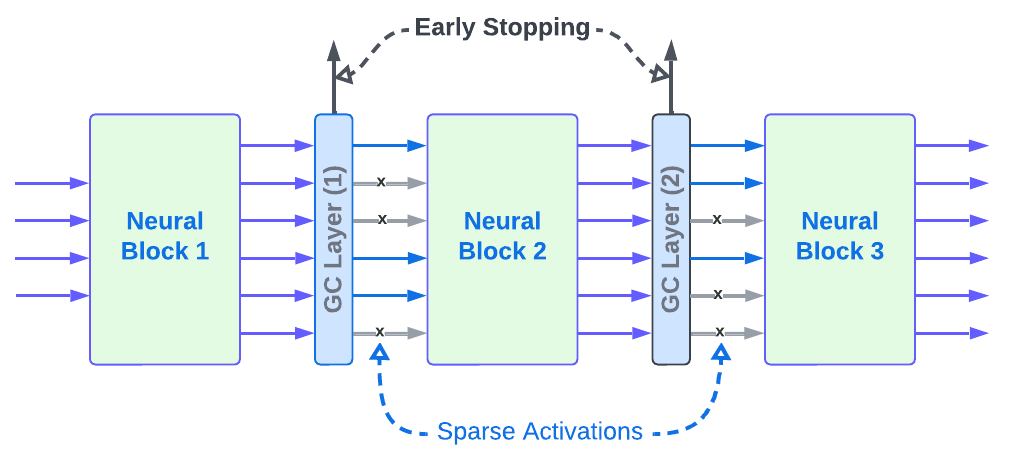}
    \caption{Two GC layers added to existing architectures to transform any
network into an efficient Always-On Gated Neural Network.}
    \label{fig:gc_4_ux}
\end{figure}

%% file: figs/dataset_arch_table.tikz
\def\arraystretch{1.25}
\begin{table*}[]
\caption{Overview of datasets and model architectures used in the study, highlighting the distinction between vision and speech datasets, the respective model architectures, implementation details, and learning rate strategies tailored to each usecase.}
\label{tab:dataset_and_arch}\begin{adjustbox}{max width=0.997\textwidth}
\begin{tabular}{l||cc|cc}
\hline\hline
                            & \multicolumn{2}{c}{\textsc{Vision}}         & \multicolumn{2}{c}{\textsc{Speech}}            \\\hline\hline
Dataset                           & \multicolumn{2}{c|}{ImageNet 2012 \cite{ILSVRC15}} & \multicolumn{2}{c}{Speech Command \cite{speechcommandsv2}}           \\
Model Architecture                & \multicolumn{2}{c|}{ResNet-152 \cite{he2016deep-resnet-152}}  & \multicolumn{2}{c}{TC-ResNet \cite{choi2019temporal-tc-resnet}}               \\
Implementation                    & \multicolumn{2}{c|}{TensorFlow Model Garden \cite{tensorflowmodelgarden2020}}  & \multicolumn{2}{c}{Streaming Keyword Spotting \cite{rybakov2020streaming-kws-streaming}} \\
Learning Rate           & \multicolumn{2}{c|}{Cosine Decay \cite{loshchilov2016sgdr-cosine-decay}(Fig. \ref{fig:Learning_Rate}, left)}  & \multicolumn{2}{c}{Piecewise Constant Decay(Fig. \ref{fig:Learning_Rate}, right)} \\\hline
Real-World Use-Case/Scenario                          & Person Detector   & Dog Detector  & 35 Keywords Detector     & 10 Keywords Detector    \\
Number of Positive Classes/labels               & 4                 & 130           & 35                    & 10                   \\
Negative Samples / Whole Dataset                & 99.6\%              & 87.0\%          & 80.0\%                  & 80.0\%                 \\
Class Weight: Positive vs Negative & 0.99 vs 0.01        & 0.8 vs 0.2      & 0.5 vs 0.5              & 0.5 vs 0.5             \\\hline\hline
\end{tabular}
\end{adjustbox}
\end{table*}



\begin{figure*}
\centering
\begin{minipage}{.47\textwidth}
  \centering
    \includegraphics[width=0.98\textwidth]{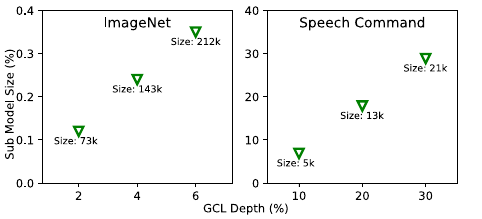}
    \captionof{figure}{Comparative Analysis of sub-model size (i.e., initial network before GC layer) relative to the placement depth of the GC layer for ImageNet and Speech Command experiments.}
    \label{fig:sub_model_size}
\end{minipage}%
\hspace*{5mm}
\begin{minipage}{.47\textwidth}
  \centering
    \includegraphics[width=0.975\textwidth]{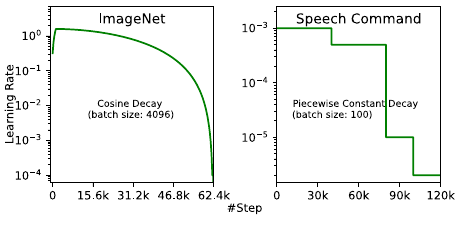}

    \captionof{figure}{Learning Rate Schedules, depicting the Cosine Decay schedule for ImageNet and the Piecewise Constant Decay schedule for speech command.}
    \label{fig:Learning_Rate}
\end{minipage}%
\end{figure*}

%% file: figs/overall_table.tikz
\setlength\tabcolsep{3 pt}
\begin{table*}[]
\caption{Comprehensive Performance Analysis: This table displays the precision, recall, gating performance and activation sparsity for person detector, dog detector, 35 keywords and 10 keywords detector models. Baseline metrics are compared against models with different depths of GC layer integration, demonstrating the impact of GC layer on model prediction performance. }
\resizebox{0.99\textwidth}{!}{
\label{tab:overall_performance}
\begin{tabular}{l||l|cccc|cccc|ccc|ccc}
\hline\hline
\multicolumn{2}{l|}{\multirow{2}{*}{}}          & \multicolumn{4}{c|}{Precision}                                                     & \multicolumn{4}{c|}{Recall}                                                        & \multicolumn{3}{c|}{Early Stopping (Gating)}                  & \multicolumn{3}{c}{Activation Sparsity}                      \\\cline{3-16}
\multicolumn{2}{l|}{}                           & Baseline           & GCL@1             & GCL@2             & GCL@3             & Baseline           & GCL@1             & GCL@2              & GCL@3              & GCL@1              & GCL@2              & GCL@3              & GCL@1              & GCL@2              & GCL@3              \\
\hline\hline
\multirow{4}{*}{\rotatebox[origin=c]{90}{\textsc{Vision}}} & \multirow{2}{*}{Person Detector}      & 99.3512     & 99.4449     & {99.5193}     & \textbf{99.5320}     & 95.2378     & 95.3911     & 95.2656     & \textbf{95.7075}     & 38.1243 & 45.5015 & \textbf{52.3367}     & \textbf{99.0943}     & 99.0435     & 98.9026      \\
                        &                                       & $\pm$0.0706 & $\pm$0.0609 & $\pm$0.0529 & \textbf{$\pm$0.0406} & $\pm$0.1288 & $\pm$0.1597 & $\pm$0.1804 & \textbf{$\pm$0.1579}   & $\pm$9.5612 & $\pm$9.6713  & \textbf{$\pm$4.2481}  & \textbf{$\pm$0.0032} & $\pm$0.0052  & $\pm$0.0048 \\
                        & \multirow{2}{*}{Dog Detector}         & 96.6936     & 96.9241     & 96.8247     & \textbf{97.0122}     & 96.2765     & 96.7232     & 96.8664     & \textbf{97.1914}     & 42.2562       & 73.1245      & \textbf{94.8752}      & {98.2470}     & 97.9511     & \textbf{98.6593}      \\
                        &                                       & $\pm$0.0401 & $\pm$0.0446 & $\pm$0.0466 & \textbf{$\pm$0.0502} & $\pm$0.0494 & $\pm$0.0388 & $\pm$0.0605 & \textbf{$\pm$0.0612} & $\pm$2.5132   & $\pm$1.7628  & \textbf{$\pm$0.9114}  & {$\pm$0.0502} & $\pm$0.0508  & \textbf{$\pm$0.0406}   \\\hline
\multirow{4}{*}{\rotatebox[origin=c]{90}{\textsc{Speech}}} & \multirow{2}{*}{35 Keywords} & 96.1344     & 96.3844     & \textbf{96.8182}     & 96.3892     & 96.0691     & 96.3679     & 96.5263     & \textbf{96.7892}     &  \multicolumn{3}{c|}{\multirow{4}{*}{\textbf{100}}}      & \textbf{88.1242}     & 82.2701     & 75.6890     \\
                        &                                       & $\pm$0.0723 & $\pm$0.0799 & \textbf{$\pm$0.0966} & $\pm$0.0791 & $\pm$0.0757 & $\pm$0.0825 & $\pm$0.0936 & \textbf{$\pm$0.0699} &            &         &     & \textbf{$\pm$0.0034} & $\pm$0.0079 & $\pm$0.0071 \\
                        & \multirow{2}{*}{10 Keywords} & 97.1314     & 97.5848     & \textbf{97.6821}     & 97.6083     & 97.0029     & 97.7129     & 97.6266     & \textbf{97.8194}     &       &          &        & \textbf{95.2541}     & 89.7053     & 85.8261     \\
                        &                                       & $\pm$0.0739 & $\pm$0.0799 & \textbf{$\pm$0.0966} & $\pm$0.0902 & $\pm$0.0720 & $\pm$0.0852 & $\pm$0.0854 & \textbf{$\pm$0.0908} &               &              &              & \textbf{$\pm$0.0055} & $\pm$0.0058 & $\pm$0.0061
\\\hline\hline

\end{tabular}
}
\end{table*}
\setlength\tabcolsep{5 pt}

%% file: figs/overall_plot.tikz
\begin{figure*}[t!]
    \centering
    \includegraphics[width=0.975\textwidth]{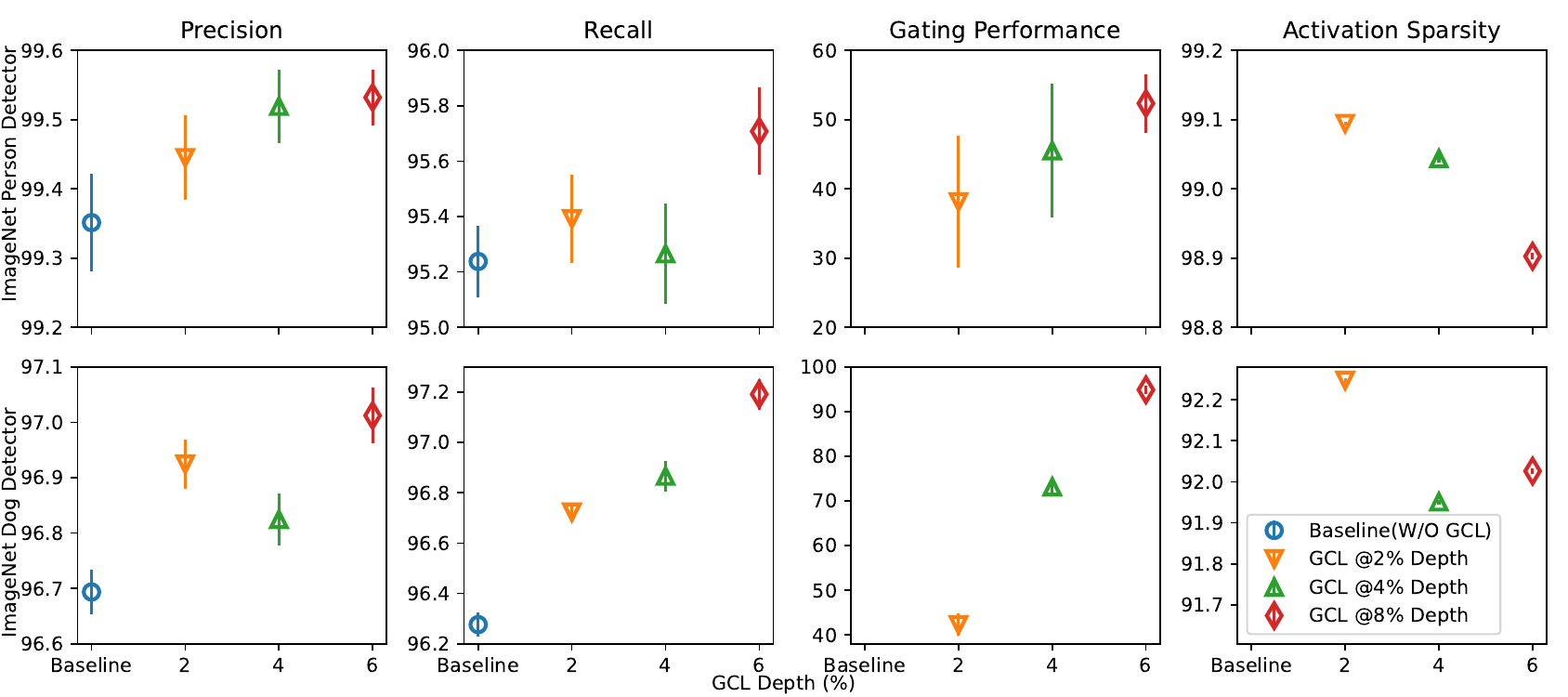}
    \caption{Performance Impact of GC layer on the ImageNet dataset. This figure illustrates the precision, recall, gating performance, and activation sparsity at various depths of GC layer integration within the network architecture, indicating the GC layer's influence on the overall model performance for the person and dog detection tasks.}
    \label{fig:imagenet_2012_overall}
\end{figure*}

\begin{figure*}[t!]
    \includegraphics[width=1.025\textwidth]{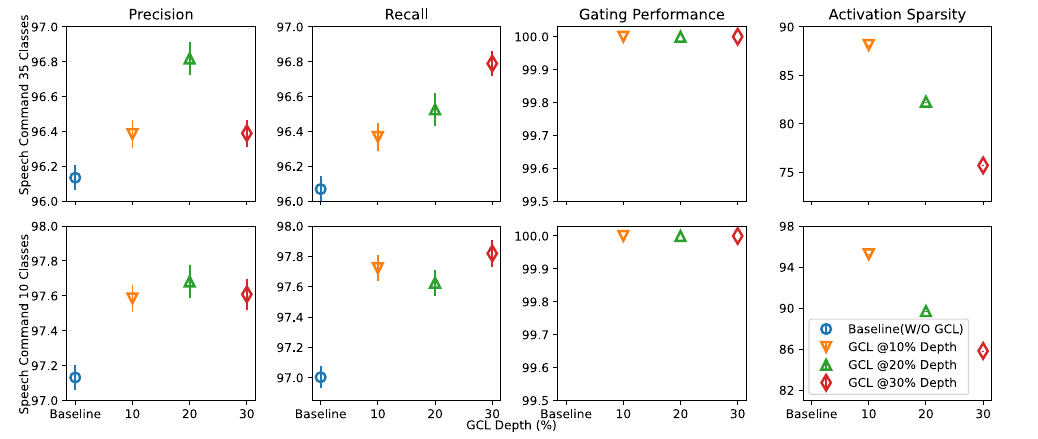}
    \caption{GC layer impact on the Speech Command Dataset. Shown here are the performance metrics of precision, recall, gating performance, and activation sparsity highlighting the effect of GC layer insertion at different depths of the network. For example, GCL@1, the first GC layer is inserted at 10\% depth of the baseline network, whereas the GCL@3, the third GC layer is inserted at 30\% depth of the baseline network.}
    \label{fig:speech_command_overall}
\end{figure*}

%% file: figs/roc_plots.tikz
\begin{figure*}[t!]
    \centering
    \includegraphics[width=0.975\textwidth]{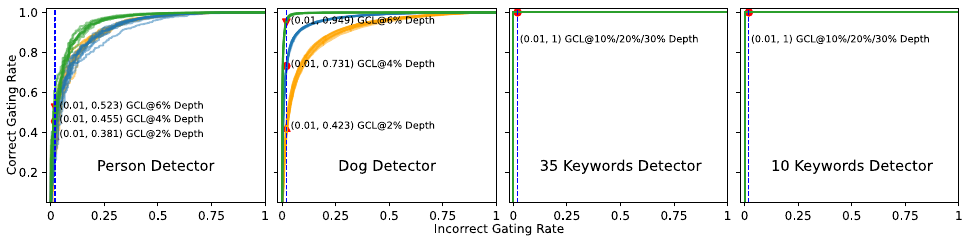}
    \caption{Gating Performance on Different Use Cases on the ImageNet and Speech Command Datasets, stratified by the depth at which the GC layer is introduced within the neural network architecture. The gating performance is quantified by the correct gating rate, with higher values indicating better selective gating of irrelevant information (i.e., early stopping of negative samples). Lastly, the correct gating results (i.e., blue lines) are highlighted by setting the incorrect gating rate (i.e., false negative rate) to a strict threshold of 1\%.}
    \label{fig:roc}
\end{figure*}

%% file: figs/power_costs.tikz
\begin{figure*}[t!]
    \centering
    \includegraphics[width=0.975\textwidth]{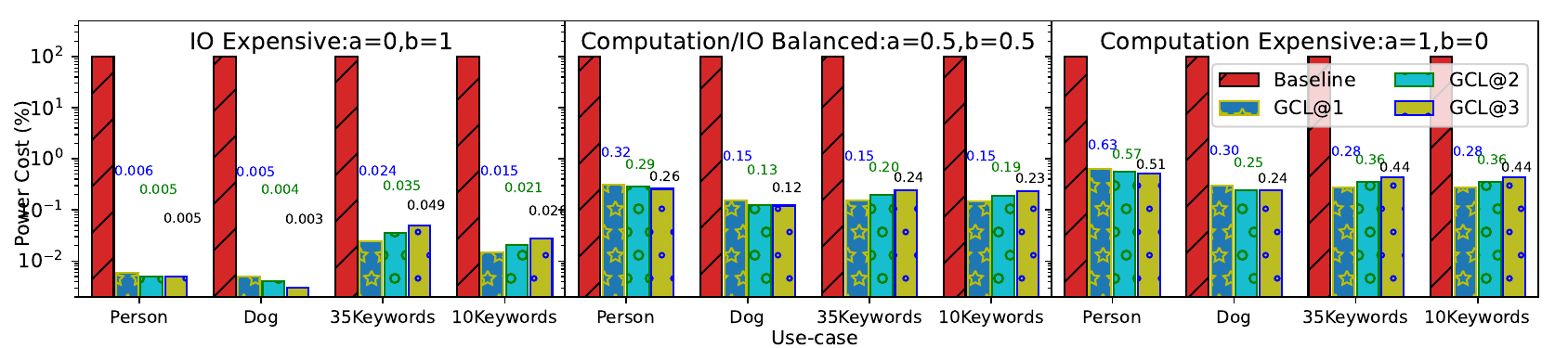}
    \caption{Power Costs Analysis of GC layer Integration Across Use-Cases. This graph quantifies the computational and transmission costs associated with GC layer depth variations in various speech and vision tasks. A lower `a' value corresponds to lower computational expense, while a lower `b' value indicates minimized data transmission requirements. Each bar represents a different configuration, comparing the baseline with three different GC layer configurations.}
    \label{fig:power_cost}
\end{figure*}

%% file: figs/vit_table.tikz
\setlength\tabcolsep{3 pt}
\begin{table*}[]
\caption{Comprehensive Performance Analysis: This table displays the precision, recall, gating performance and activation sparsity for the person, dog and cat detection task. Baseline metrics are compared against models with different depths of GC layer integration and by varying the early stopping weight ($\alpha$), demonstrating the impact of GC layer on model prediction performance. }
\resizebox{1.012\textwidth}{!}{
\label{tab:overall_vit_performance}
\begin{tabular}{l||c|c|cccc|c|cccc|cccc|cccc}
\hline\hline
&GCL&        \multicolumn{5}{c|}{Precision}                                                     & \multicolumn{5}{c|}{Recall}                                                        & \multicolumn{4}{c|}{Early Stopping (Gating)}                  & \multicolumn{4}{c}{Activation Sparsity}                      \\\cline{3-20}
&\makecell{$\alpha$}    & \multicolumn{1}{c}{Baseline}            & GCL@1             & GCL@2             & GCL@3      & GCL@4            & \multicolumn{1}{c}{Baseline}               & GCL@1             & GCL@2              & GCL@3         & GCL@4     & GCL@1              & GCL@2              & GCL@3      & GCL@4        & GCL@1              & GCL@2              & GCL@3        & GCL@4      \\
\hline\hline
\multirow{8}{*}{\rotatebox[origin=c]{90}{\textsc{\makecell{ViT-L/16: Person,\\ dog and cat Detector}}}} & \multirow{2}{*}{$0.1$}      & \multirow{8}{*}{\makecell{89.448\\$\pm$0.212}}   &  \textbf{90.906}    & 89.908     &{90.163}     & 90.156     &  \multirow{8}{*}{\makecell{90.923\\$\pm$0.119}}      & 91.414     & 91.297 & \textbf{91.468}  & 91.460      & 44.391 & 51.902 & 58.482 & \textbf{59.481}     & 89.726     & 90.461     & 90.555 & \textbf{90.758}      \\
                        &                                       &  & \textbf{$\pm$0.374} & $\pm$0.726 & {$\pm$0.289} & $\pm$0.330 & & $\pm$0.267 & $\pm$0.516 & \textbf{$\pm$0.183}  & $\pm$0.241 & $\pm$1.823 & $\pm$1.462  & $\pm$1.517 & \textbf{$\pm$2.653}  & {$\pm$0.162} & $\pm$0.059  & $\pm$0.025 & \textbf{$\pm$0.077} 
\\\cline{2-2}\cline{4-7}\cline{9-20}
 & \multirow{2}{*}{$0.3$}      &   & 90.896  & 89.997  & 90.233  & \textbf{90.256}  &   & 91.403  & 91.397  & \textbf{91.481}  & 91.469  & 49.219  & 55.942  & 62.823  & \textbf{69.881}  & 89.926  & 90.191  & 90.225  & \textbf{90.651}      \\
                        &                                       &   & $\pm$0.314  & $\pm$0.291  & $\pm$0.219  & \textbf{$\pm$0.302}  &   & $\pm$0.169  & $\pm$0.316  & \textbf{$\pm$0.389}  & $\pm$0.241  & $\pm$1.837  & $\pm$1.629  & $\pm$1.812  & \textbf{$\pm$2.153}  & $\pm$0.092  & $\pm$0.092  & $\pm$0.071  & \textbf{$\pm$0.071}
\\\cline{2-2}\cline{4-7}\cline{9-20}
 & \multirow{2}{*}{$0.5$}      &   & 90.162  & 89.981  & \textbf{90.321}  & 90.256  &   & 91.215  & 91.217  & 91.318  & \textbf{91.362}  & 51.951  & 61.321  & 68.822  & \textbf{75.814}   & 89.962  & \textbf{90.621}  & 90.515  & 90.583      \\
                        &                                       &  & $\pm$0.274  & $\pm$0.314  & \textbf{$\pm$0.291}  & $\pm$0.293  &   & $\pm$0.226  & $\pm$0.213  & $\pm$0.238  & \textbf{$\pm$0.244}  & $\pm$1.831  & $\pm$1.622  & $\pm$1.712  & \textbf{$\pm$2.032}  & $\pm$0.069  & \textbf{$\pm$0.089}  & $\pm$0.055  & $\pm$0.087 
\\\cline{2-2}\cline{4-7}\cline{9-20}
& \multirow{2}{*}{$0.7$}      &   & 89.961  & \textbf{90.522}  & 90.469  & 90.456  &   & 91.149  & 91.295  & 91.388  & \textbf{91.426}  & 56.991  & 68.925  & 78.829  & \textbf{85.981}   & 89.996  & 90.216  & \textbf{90.325}  & 90.251      \\
                        &                                       &   & $\pm$0.274  & \textbf{$\pm$0.296}  & $\pm$0.291  & $\pm$0.255  &   & $\pm$0.317  & $\pm$0.287  & $\pm$0.258  & \textbf{$\pm$0.249}  &  $\pm$1.869  & $\pm$1.961  & $\pm$1.72  & \textbf{$\pm$2.152}   & $\pm$0.062  & $\pm$0.051  & \textbf{$\pm$0.052}  & $\pm$0.049
\\\hline\hline
\end{tabular}
}
\end{table*}
\setlength\tabcolsep{5 pt}

%% file: figs/vit_imagenet_results.tex
\begin{figure}[t!]
    \centering
    \includegraphics[width=0.495\textwidth]{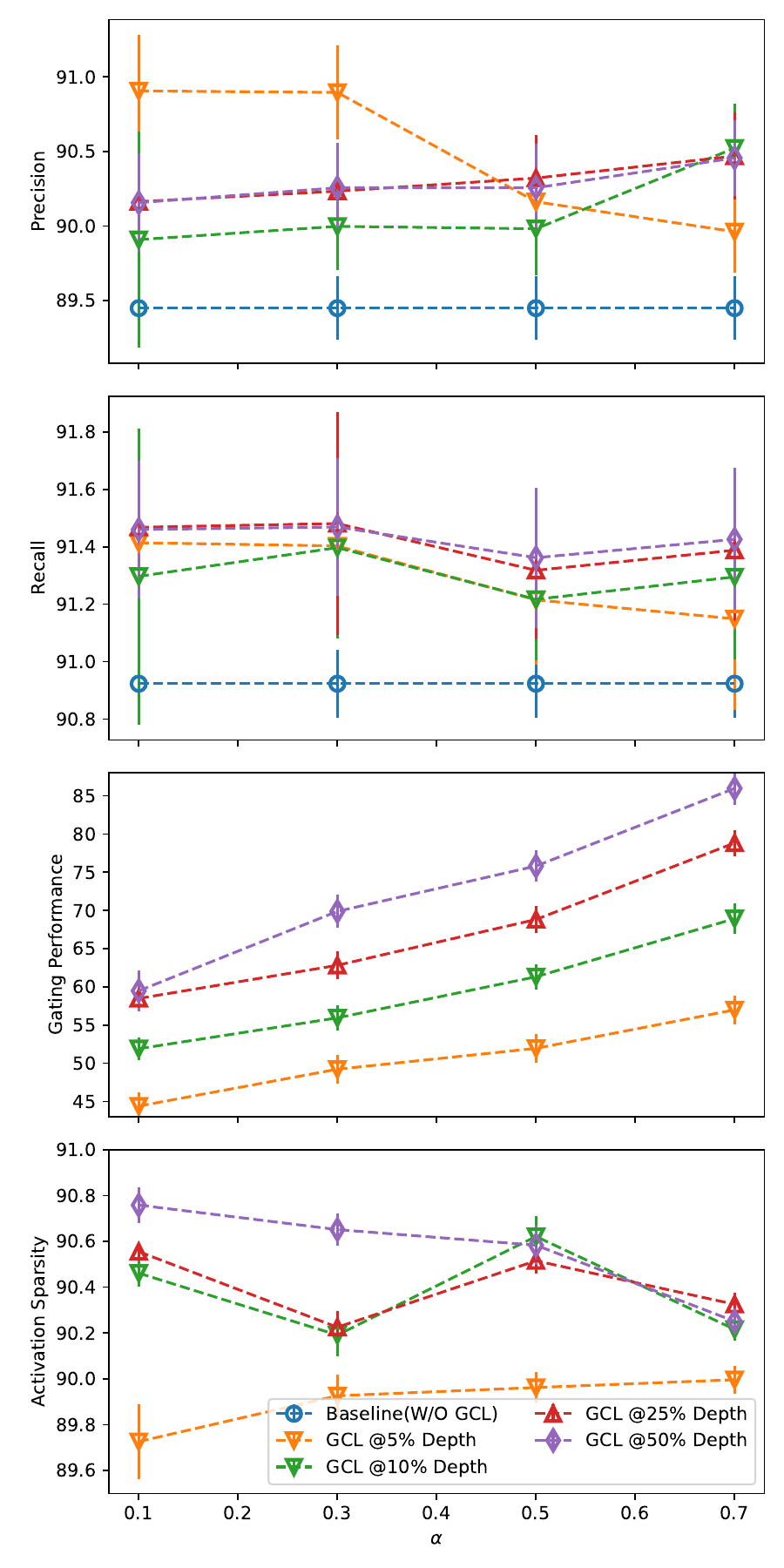}
    \caption{Impact of GC layer Depth and Early Stopping Weight on ViT-L/16 Performance. This figure demonstrates the effect of varying the GC layer depth and the early stopping weight (referred to as $\alpha$) on the performance of ViT-L/16 architecture. The experiment measures the precision, recall, gating performance, and activation sparsity at various depths of GC layer integration, each tested with different $\alpha$ values as illustrated in figure. }
    \label{fig:vit-alpha-performance-pdf}
\end{figure}

\begin{figure*}[t!]
    \centering
    \includegraphics[width=0.975\textwidth]{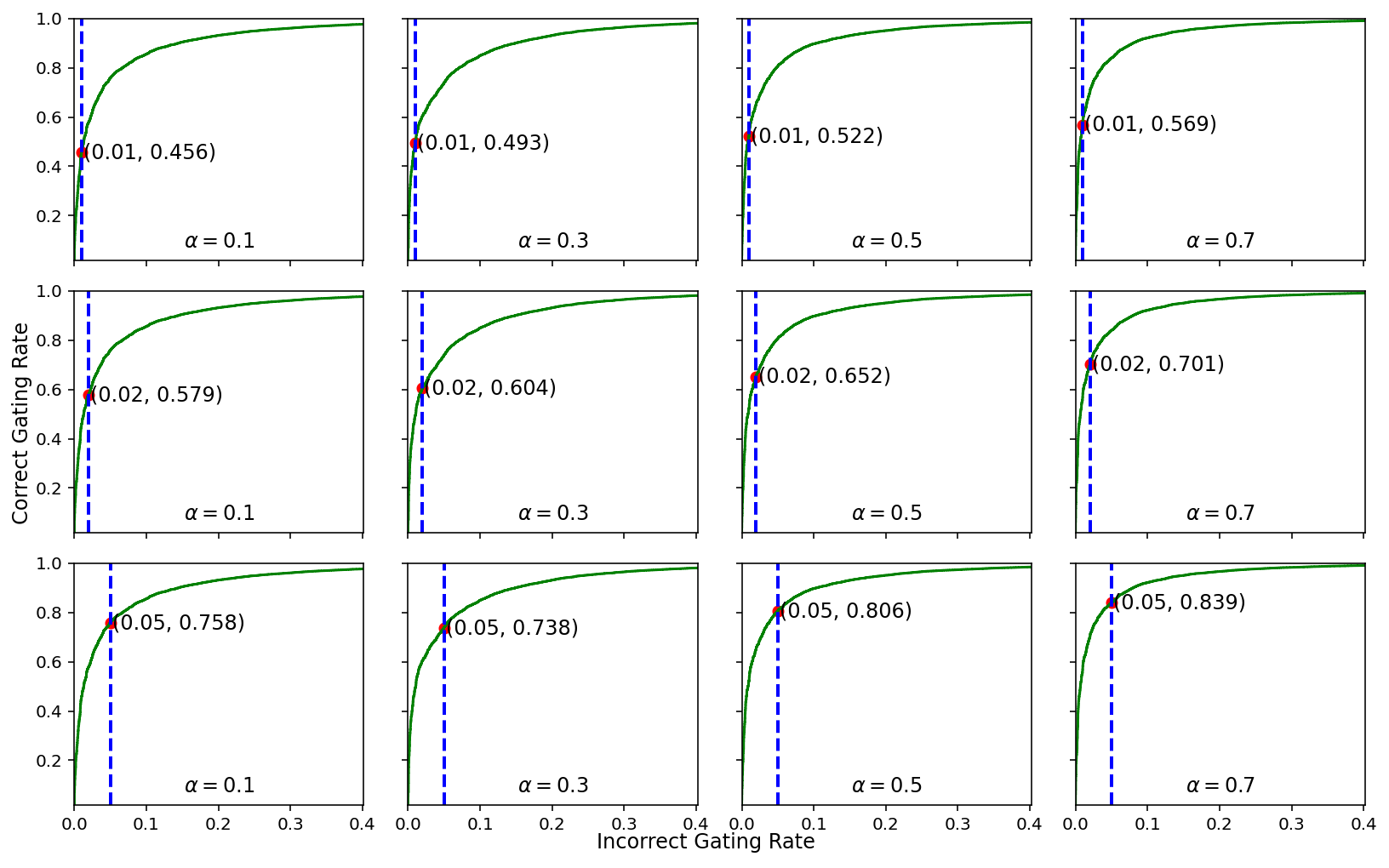}
    \caption{Impact of Adjusting Incorrect Gating Rate on Vit-L/16. This figure shows the relationship between varying incorrect gating rates for different values of the early stopping weight, $\alpha$. The graph highlights how changing the threshold for incorrect gating (indicated by the blue dashed lines) affect the model's ability to correctly gate computations as shown by the green curves.}
    \label{fig:vit-alpha-performance}
\end{figure*}